\def\year{2021}\relax
\documentclass[letterpaper]{article} 
\usepackage{aaai21}  
\usepackage{times}  
\usepackage{helvet} 
\usepackage{courier}  
\usepackage[hyphens]{url}  
\usepackage{graphicx} 
\usepackage{natbib}  
\usepackage{caption} 
\usepackage{amsmath}
\usepackage{booktabs}
\usepackage{xcolor}

\usepackage[normalem]{ulem}

\newcommand\datasetName{MultiTalk Dataset}
\newcommand\modelName{Lookahead Model}

\newcommand\emoModelName{Emotion-Aware Model}

\DeclareMathOperator*{\argmax}{argmax}

\title{MultiTalk: A Highly-Branching Dialog Testbed for Diverse Conversations}
\author{
    Yao Dou,$^*$ Maxwell Forbes,$^{*\dagger}$ Ari Holtzman,$^*$ Yejin Choi$^{*\dagger}$ \\
}
\affiliations {
    $^*$University of Washington, $^\dagger$Allen Institute for AI \\
    \{douy, mbforbes, ahai, yejin\}@cs.washington.edu
}

\begin{document}
\maketitle

\begin{abstract}

We study conversational dialog in which there are many possible responses to a given history.
We present the \datasetName, a corpus of over 320,000 sentences of written conversational dialog that balances a high branching factor (10) with several conversation turns (6) through selective branch continuation.
We make multiple contributions to study dialog generation in the highly branching setting.
In order to evaluate a diverse set of generations, we propose a simple scoring algorithm, based on bipartite graph matching, to optimally incorporate a set of diverse references.
We study multiple language generation tasks at different levels of predictive conversation depth, using textual attributes induced automatically from pretrained classifiers.
Our culminating task is a challenging theory of mind problem, a controllable generation task which requires reasoning about the expected reaction of the listener.

\end{abstract}

\section{Introduction}

A range of work in artificial intelligence has studied natural language generation in conversational dialog \cite{adiwardana2020towards, zhang2018personalizing, serban2016building}. 
Dialogs provide rich scenarios for understanding social interaction and real-world knowledge.
But within a dialog, the question often arises: What if a character had said something else?
Such counterfactuals can be powerful tools to learn from, and children tend to naturally explore such possibilities naturally by asking ``what if'' questions \cite{bates1976language,au1992counterfactual}.

Inspired by these aims, we introduce the large-scale \datasetName~of highly-branching dialogs, where each character gives several (10) responses to the full conversation history at that point.
The high branching factor leads to many different conversational paths, which we manage by continuing the dialog for only a subset of responses.
In total, the dataset contains over 320,000 sentences across 120 conversation trees, where each tree contains a unique prompt and pair of characters that alternate responses.

\begin{figure}[t]

\begin{center}
\includegraphics[width=0.9\linewidth]{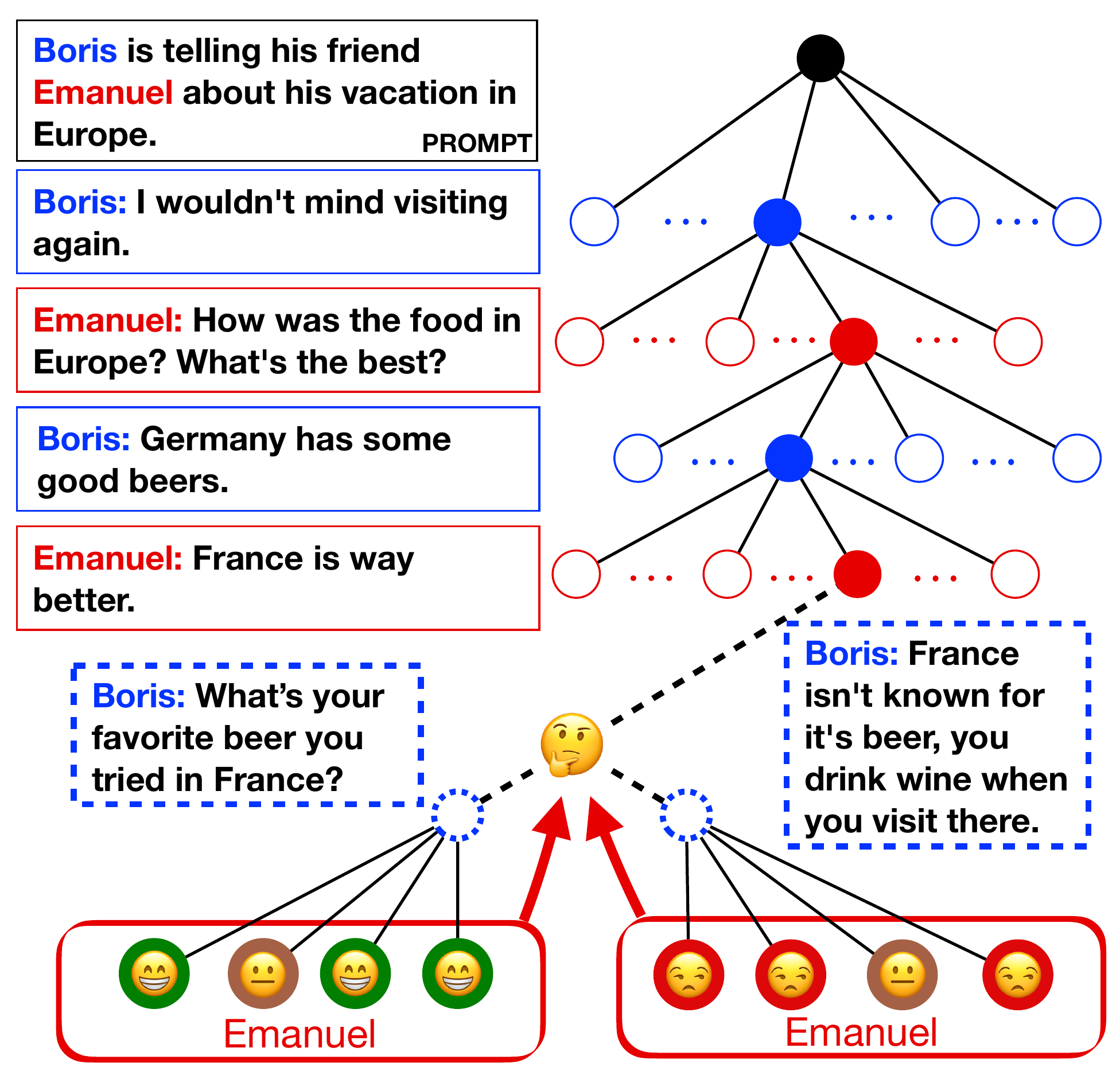}
\end{center}

\caption{
We introduce the highly branching \textit{MultiTalk} dialog dataset, with over 320k sentences.
We study several tasks in this new domain, including a \textit{theory of mind} task, where a model estimates how listeners will react to its output.
}

\label{fig:figure1}
\end{figure}
\begin{figure*}[th!]

\begin{center}
\includegraphics[width=0.99\linewidth]{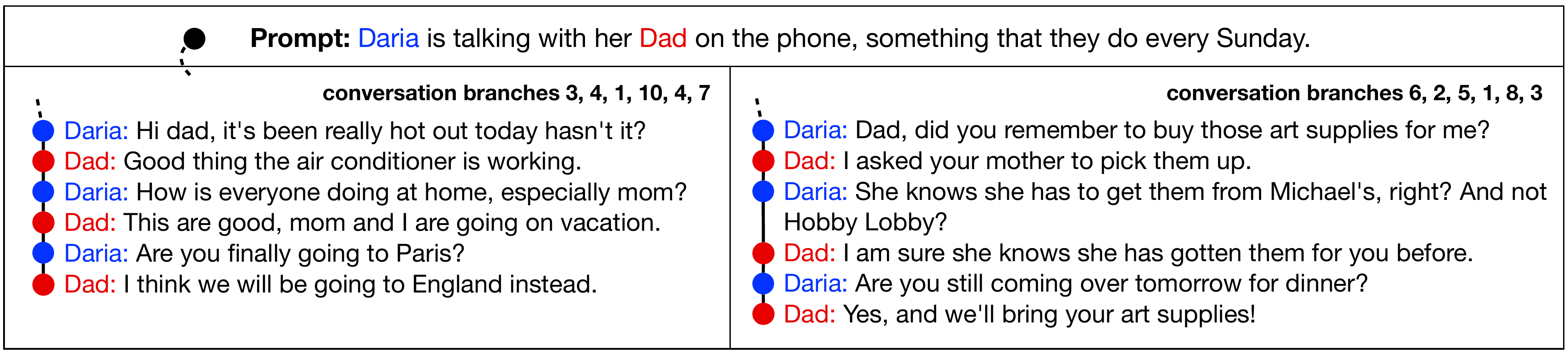}
\end{center}

\caption{
With a high (10) branching factor, a single prompt in the \datasetName~leads to a wide variety of conversations.
Pictured here are two example paths from the training set that follow from a single prompt.
}

\label{fig:datasetExample}
\end{figure*}

To study model performance on this dataset, we introduce a new scoring technique that accounts for the diversity of responses.
Naively applying an evaluation metric like BLEU \cite{papineni2002bleu} would assign high points to a model that repeatedly outputs a single well-scoring response.
To account for this, we propose a simple scoring algorithm based on weighted bipartite graph matching.
This technique dynamically matches candidates to references, assigning the maximum score to candidates while ensuring each reference is used only once.
The technique requires no changes to models, can accept any individual scoring function to compare a candidate and reference (e.g., BLEU, BERT-Score \cite{bert-score}, etc.), and is solvable in polynomial time thanks to the Kuhn-Munkres algorithm \cite{kuhn1955hungarian}.

We also use the \datasetName~as an opportunity to explore models that attempt to optimize for particular dialog paths.
First, we provide benchmark performances for a diverse response generation task, using \datasetName~as a testbed.
Then, we propose two tasks to study controllable generation on a dialog tree using induced textual attributes.
The second of these tasks emulates a \textit{theory of mind} setting, where models must induce attribute-valued replies from a conversation partner who lacks explicit attribute knowledge.

\section{\datasetName}
\begin{table}[tp]
\small
\centering
\scalebox{1}{%
\begin{tabular}{@{}lr@{}}
\toprule
Branching factor ($b$)  &  10\\
Continuation factor ($c$)  &  3\\
Max response depth ($d$) &  6\\ 
Total prompts ($u_0$) & 120  \\ 
Total sentences  &  320,804 \\ 
Avg. sentences per prompt & 2673\\
Avg. sentence length & 8.5 \\
\bottomrule
\end{tabular}}
\caption{ Statistics of our proposed \datasetName.
}
\label{tab:dataset_stats}
\end{table}
We collect and release\footnote{\url{https://uwnlp.github.io/multitalk/}} a large dataset of highly-branching written conversations.
The dataset contains 320,804 individual responses in a conversation tree.
Figure \ref{fig:datasetExample} shows two example conversation paths from the same prompt in the dataset. 
Summary statistics of the dataset are given in Table~\ref{tab:dataset_stats}.
Each dialog begins with a pre-specified scenario that establishes two characters and a setting.
From this scenario, the conversation branches and deepens.
Each character provides $b=10$ potential responses or \textit{branches} to the state of the conversation so far.
From a randomly-chosen continuation subset $c=3$ of those responses, the conversation continues with the next character.
This continues to a maximum depth of $d=6$ conversation turns.
The following sections describe both the scenarios and conversation turns.

\subsection{Scenarios: Prompts and Characters}

We manually write $n=120$ prompts.
The prompts are motivated by everyday situations.
We draw inspiration from a list of conversation questions for English language learners.\footnote{\url{http://iteslj.org/questions/}}
Each prompt is a single sentence.

The characters names are manually chosen using random name generators.
We begin from a list of gender-balanced names, and assign a gender pronoun for each character.
The list contains representations from names common among different cultural groups.
The base prompt is combined with the two characters in order to form the scenario.
We adjust the wording of scenarios to match the genders (e.g., \textit{girlfriend} or \textit{boyfriend}).
Relationships are formed from all candidate gender pairings.
Some prompt/name pairings are purposefully left gender ambiguous.

\subsection{Conversation Turns}



At each turn, an annotator is provided with the full conversation context so far, and writes all $d = 10$ responses for the current speaker.
A single annotator completes each step to minimize cognitive load; rather than read and characterise a partial set of existing responses, an annotator must only reason about the set of responses they will write.
Annotators are free to mix both surface and semantic diversity \cite{tevet2020evaluating}.
We perform manual quality control by checking a sample of work from each annotator and conversation tree during each round of annotation. 

\subsection{Dataset Quality and Utility}


\begin{table}[tp]
\centering
\resizebox{\columnwidth}{!}
{\begin{tabular}{@{}lccc@{}}
\toprule
\textbf{Model} & \textbf{Diversity} & \multicolumn{1}{p{3cm}}{\centering \textbf{Relevance}}  & \multicolumn{1}{p{3cm}}{\centering \textbf{Fluency} } \\ \midrule
GPT-2 XL   &    65.0\%   &  98.8\%   &    99.0\% \\ 
Dialog Tree Model                                                                  &    85.6\%    &  100.0\% &   99.0\% \\
\bottomrule
\end{tabular}}
\caption{
How often the MultiTalk dataset (reference) annotation is rated as superior to top performing models by humans along three dimensions.
Each comparison pits references against 10 model generations; 500 compairsons per model. \textit{Takeaway:} the MultiTalk Dataset has significantly more diverse responses than top performing models, and overwhelmingly better relevance and fluency.
}
\label{tab:task0_human_eval_vs}
\end{table}
\begin{figure}[t]

\begin{center}
\includegraphics[width=0.99\linewidth]{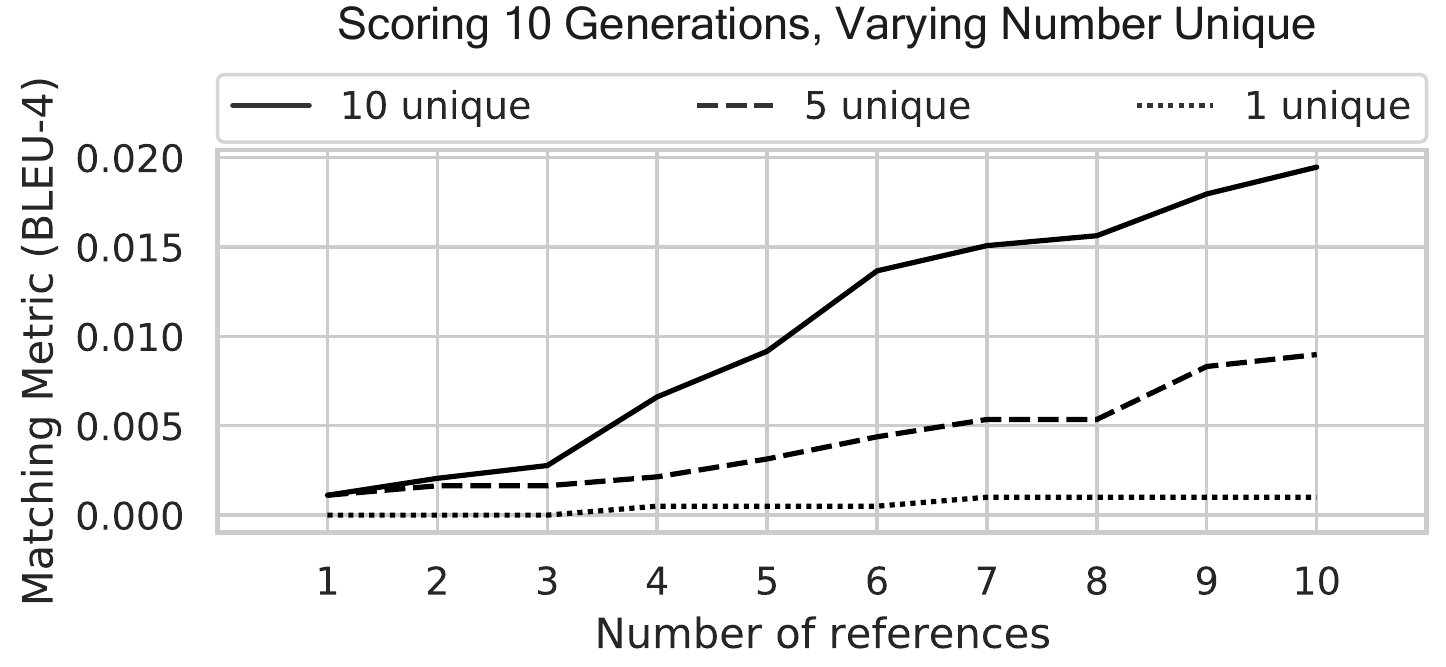}
\end{center}

\caption{
Varying the number of references in order to distinguish diverse generations.
\textit{Takeaway:} With only 1 reference, metrics have difficulty distinguising between more and less diverse model outputs. With 10 references (our proposed dataset), models can be clearly stratified by diversity.
}

\label{fig:metric_vs_references}
\end{figure}

We briefly present two measurements of the dataset to asses its quality and utility.
While these measurements use models and techniques described in later sections of the paper, we present them here as they primarily illustrate properties of the dataset.

Table \ref{tab:task0_human_eval_vs} assesses dataset \emph{quality}. It compares top performing models (\S\ref{sec:model}) against the dataset itself in a blind human evaluation.
Models are tasked to perform \textit{Diverse Generation} (\S\ref{sec:tasks})---in other words, generate 10 responses to a context, just like the dataset.
We observe the dataset is consistently more diverse, and resoundingly more relevant and fluent, than state-of-the-art neural language models.

Figure \ref{fig:metric_vs_references} demonstrates an aspect of the dataset's \emph{utility}.
This graph controls the number of unique generations from a model (different lines), and scores them against a varying number of references ($x$-axis) with a matching algorithm (\S\ref{sec:matching}).
We observe that having many references allows evaluations on the MultiTalk dataset to distinguish between more and less-diverse model outputs.
\section{Scoring Algorithm}
\label{sec:matching}

\noindent
\textbf{Evaluating Diverse References} In dialogs with a high branching factor, multiple references are available for any candidate generation.
How should we evaluate generations based on those references?

\begin{figure}[t]

\begin{center}
\includegraphics[width=0.7\linewidth]{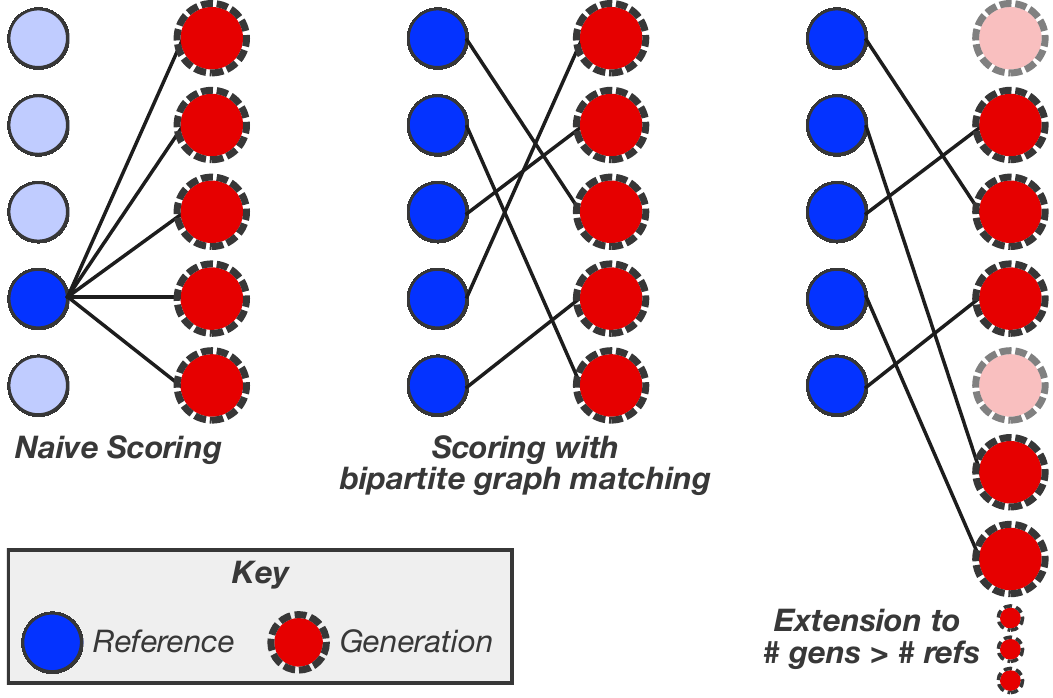}
\includegraphics[width=0.8\linewidth]{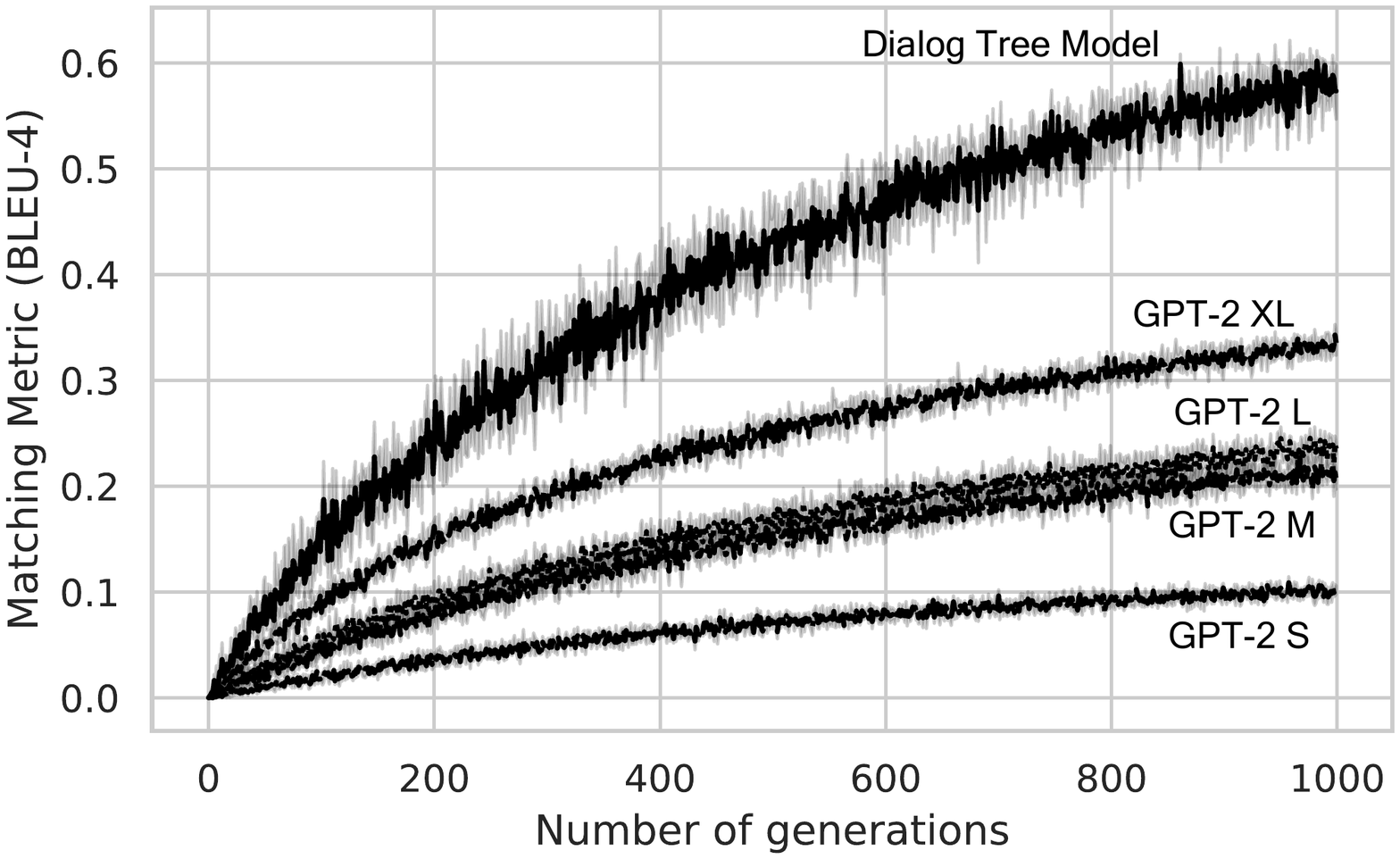}
\end{center}
\caption{
\textit{Top:} By casting evaluation as a matching problem on a weighted bipartite graph, we can harness any pairwise scoring metric to evaluate a set of generations against a potentially diverse set of references.
This technique extends naturally to a greater (or fewer) number of generations.
\textit{Bottom:} Match scoring against a diverse set of references with a tough metric (BLEU-4) smoothly scaling to significantly more generations.
}

\label{fig:matching}
\end{figure}

A naive approach is to simply run a metric like BLEU out-of-the-box.
Figure \ref{fig:matching} (top, left) illustrates an issue with this approach.
Each generation can use the same reference to receive their score.
This means that a model could achieve a high score simply by producing a single high-scoring response many times.
This is not an issue in semantically-constrained generation tasks, as in machine translation, because all references attempt to say the same thing.
However, for a dataset like ours where responses are intentionally diverse, this behavior is problematic.

\vspace*{2mm}
\noindent
\textbf{Assignment Problem}
To address this issue, we can pose this task as an assignment problem.
First, we add our set of references $R$ and generations $G$ to a graph as vertices.
We can assign an edge $(r,g)$ between each reference $r \in R$ and each generation $g \in G$ with a weight of $s(r, g)$ based on a scoring function of our choosing.
For example, we could use the BLEU score between $r$ and $g$ edge weight between them.
We now have a weighted bipartite graph, and we can find the assignment

\begin{equation}
    f^* = \argmax_{f \in \mathcal{F}}  \sum_{r \in R} s(r, f(r))
\end{equation}

where $\mathcal{F}$ is the set of injective functions $f: R \rightarrow G$.
The matching $f^*$ produces the highest sum of edges that use each reference exactly once (Figure \ref{fig:matching} top, center).

The Kuhn-Munkres algorithm, also known as The Hungarian Algorithm, is a polynomial time solution to that assignment problem \cite{kuhn1955hungarian}.
While the algorithm is traditionally used to assign a matching with minimum cost, we may simply invert scores beforehand and use it as usual.\footnote{We use the \textit{linear sum assignment} implementation in SciPy (\url{https://scipy.org/}).}.

This technique has the advantage of being generous in its scoring of generations:
each generation is allowed to match to any of the references, and this assignment changes optimally as new generations are introduced.
Furthermore, the algorithm is fair, in that each reference is only allowed one usage.
And lastly, the underlying models need not consider this process at all, nor specify any assignments themselves.

\vspace*{2mm}
\noindent
\textbf{More Generations}
When $|G| = |R|$, an equal number of generations and references, $f^*$ will be a bijection.
But in practice, matching to even 10 references (e.g., in our dataset) can produce low scores depending on the metric used.\footnote{Intuitively, note that the space of potential reasonable responses in a conversational dialog is vast, certainly $\gg 10$.}
Fortunately, because $F$ is defined to be the set of \textit{injective} functions, this approach extends naturally to scoring more generations $|G| > |R|$ (see Figure \ref{fig:matching} top, right).
The matching $f^*$ will still use each reference $r \in R$ once, and will simply  pick the set of highest scoring $|R|$ generations from $G$.

\section{Tasks}
\label{sec:tasks}

We use the \datasetName~to pose three tasks that we study in upcoming sections.

\vspace*{2mm}
\noindent
\textbf{Diverse Generation}
The first task is a straightforward language generation task applied to the proposed dataset.
A model is given a conversation history $(u_0,...,u_{i-1})$ and must produce $n$ candidate generations for $u_i$.
These generations are compared against the set of references for $u_i$ using our proposed matching metric to account for diversity.

\vspace*{2mm}
\noindent
\textbf{Emotion-Conditioned Generation}
For the second task, we first train an emotion classifier (details in \S\ref{sec:emotion-classification}) to label each response in the dataset with an emotion $e \in E$.
Then, the task is to generate $|E|$ responses for $u_i$ given the conversation history and each emotion $(u_0,...,u_{i-1}, e)$.
Models must succeed according to both the emotion classifier, which rates conformance to the provided emotions, as well as an \textit{evaluative} language model fine-tuned on the dataset, which judges the coherence of the generation.

\vspace*{2mm}
\noindent
\textbf{Theory of Mind Generation}
The final task we term a \textit{theory of mind task}.
In this task, a model is paired with a conversation partner: a fine-tuned language model that has no knowledge of emotions.
A model must again produce $|E|$ responses for $u_i$ given the conversation history and each emotion  $(u_0,...,u_{i-1}, e)$.
However, in this setting, it is not the model's generation $u_i$, but its conversation partner's reply $u_{i+1}$, that must match the provided emotion $e$. 
\section{Models}
\label{sec:model}
We present adaptations of pretrained language models to the three tasks we consider.
In practice, we use the GPT2 architecture from \cite{radford2019language}, but our objectives are agnostic to the particular language model used.



\subsection{Dialog Tree Model}
\label{sec:dialog-tree-model}

\begin{figure}[t]

\begin{center}
\includegraphics[width=0.80\linewidth]{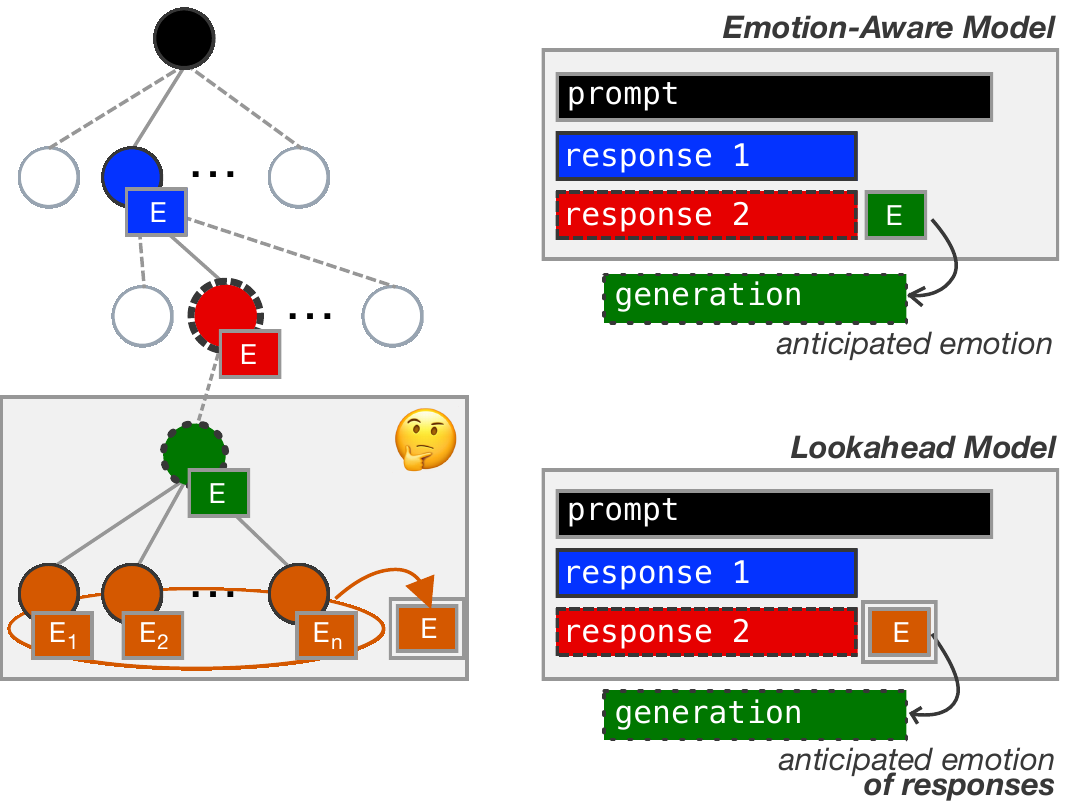}
\end{center}

\caption{
We study an Emotion-Aware Model, which anticipates the emotion of the next response, as well as a \modelName, which anticipates the emotions of \textit{responses to} its own generations.
}

\label{fig:model}
\end{figure}

For the diverse generation task, we perform a typical language model fine-tuning, with some changes to take into account the tree structure of the dialog data.
We concatenate the prompt and full conversation history together for the full path leading to each utterance in the dataset $\textbf{u} = [u_0; u_1; ...; u_i]$.
Then, we fine-tune a pretrained language model using the standard language modeling objective

\begin{equation}
    p(\textbf{u}) = p(w_1 ... w_T) = \prod_{t=1}^{T}p(w_t|w_{<t}).
    \label{con_prob}
\end{equation}

However, given the structure of our data, earlier utterances (e.g., $u_1$) are seen exponentially more frequently than later utterances (e.g., $u_6$).
To prevent biasing the language model to utterances higher in the dialog tree, we compute loss for the model only for
tokens in the final utterance:

$$
\text{loss}(w_t) =
\left\{
	\begin{array}{ll}
		- \log \left( p(w_{t}|w_{<t}) \right) & \mbox{if } w_t \in u_i \\
		0  & \mbox{otherwise}
	\end{array}
\right.
$$




\subsection{Emotion-Aware Model}
\label{sec:emotion-aware-model}

For the emotion-conditioned generation task, we develop a model that accounts for emotion state.
The Emotion-Aware Model uses the same basic structure as the Tree Dialog Model, including only computing loss on tokens for the target utterance $u_i$ and not the history $u_{<i}$. 
However, it is also given the emotion $e$ of the latest response, which it conditions on when computing probabilities


\begin{equation}
    p(w_t | w_{<t}, e).
    \label{emotion_prob}
\end{equation} 

A sketch of this model is given in Figure \ref{fig:model} (top right).

\subsection{Lookahead Model}
\label{sec:lookahead-model}

Finally, we present a lookahead model that addresses the theory of mind task.
Its training objective models responses $u_{i}$ that \textit{result} in future responses $u_{i+1}$ with the emotion $e$.

Intuitively, a model that would like to account for the emotion $e$ of all responses below it needs to consider the emotions of all children. 
If we were interested in getting the emotional state of the complete tree below the utterance $u_i$, we would compute the emotion of all subtrees as a \textit{depth-weighted estimate}, recursively defined as:

\begin{equation}
    d(u_i) = \frac{1}{|c(u_i)|} \sum_{u_{i+1} \in c(u_i)}{e(u_{i+1})} + \gamma d(u_{i+1})
\end{equation}

where $c(x)$ are the children of $x$, $e(x)$ is the (estimated) emotion of $x$, and $\gamma$ is a discount factor either down or up-weighting utterances deeper in the conversation subtree.
For practical purposes, we restrict our modeling to strictly optimize for the theory of mind task, and set $\gamma = 0$ to account only for the emotion of a response's immediate children, and choose the emotion $e$ that has the highest average in $d(u_i)$.

The per-token probability model for the Lookahead Model is then

\begin{equation}
    p(w_t | w_{<t}, d(u_i)), \quad d(u_i) \in E
    \label{lookahead_prob}
\end{equation} 

with $d(u_i)$ now the estimated emotion with the largest magnitude among the children of $u_i$.






\section{Experiments and Results}
\subsection{Emotion Recognition}
\label{sec:emotion-classification}

\begin{figure}[t]

\begin{center}
\includegraphics[width=0.99\linewidth]{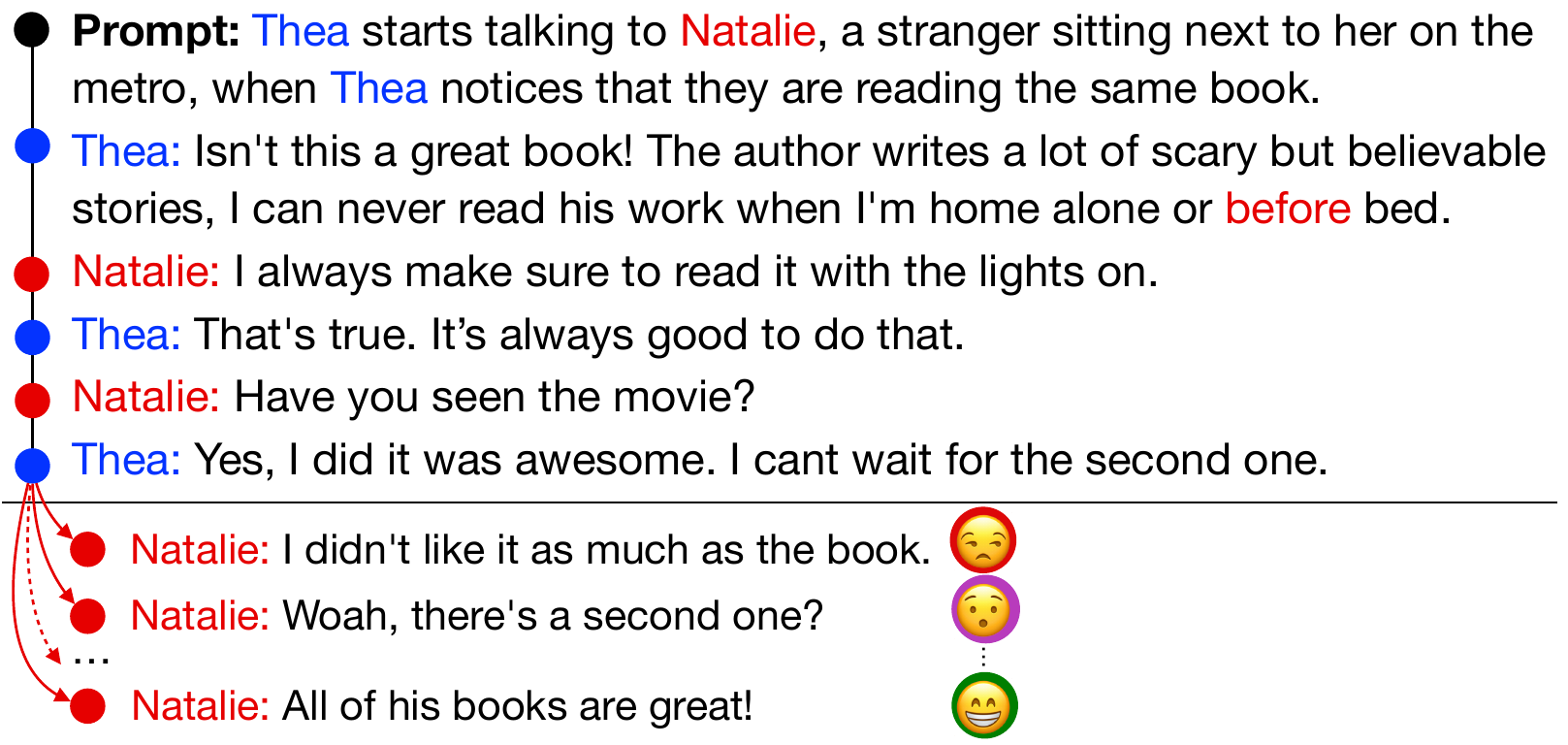}
\end{center}

\caption{
We induce emotion labels on the \datasetName~with a classifier trained on external emotion datasets.
Here, we observe a diverse distribution of emotions for an example set of responses.
}

\label{fig:emotionVariance}
\end{figure}

\noindent
\textbf{Classifier}
We train BERT-Large \citep{devlin2018bert} as an emotion classifier, fine-tuning it on multiple emotion-labeled corpora.
We use two datassets from the EmotionX 2019 Challenge \citep{shmueli2019socialnlp} and part of the \textsc{EmpatheticDialogues} dataset \citep{rashkin2018towards}.
We use the seven emotions from the EmotionX datasets: joy, sadness, fear, anger, surprise, disgust, and neutral.

The accuracy of the trained emotion classifier is 82.6\%.
Figure \ref{fig:emotionVariance} three classified responses from the dataset.

\vspace*{2mm}
\noindent
\textbf{Label Distribution}
Because our dataset is collected without emotion labels, the classified emotion frequencies end up naturally imbalanced.
While this has the advantage of potentially reflecting a more realistic distribution of conversational emotions \citep{poria2018meld}, it creates an additional challenge during modeling. We thus also consider oversampling to smooth this distribution.

\vspace*{2mm}
\noindent
\textbf{Oversampling}
For the purposes of the emotion-conditioned tasks, we create augmented versions of our dataset with two techniques.
First, we add German and Russian back-translations of sentences with rare emotions (e.g., \textit{disgust}) using \cite{ng2019facebook,ott2019fairseq}.
Second, we sample utterances from each emotion equally to create an emotion-balanced version of our dataset.
We refer to this version of the dataset as \textit{oversampled} when training models with it.

\subsection{Training and Inference}

We substitute all characters names with the special tokens {\small{\texttt{[speaker1]}}} and {\small{\texttt{[speaker2]}}}.
We fine-tune GPT-2 M (345M params.) \cite{radford2019language} on all training objectives (\S \ref{sec:model}).\footnote{GPT-2 Medium is the largest model we are able to train with our available resources.} At inference time, we sample from all models using top-p sampling with $p=0.9$ \citep{holtzman2019curious}.



\subsection{Task 1: Diverse Generation}

This task is a ``standard'' dialog response generation task, but where a model must produce a diverse set of responses to each provided context to be evaluated against 10 references.
Our primary model of study is the Dialog Tree Model (\S\ref{sec:dialog-tree-model}).



\vspace*{2mm}
\noindent
\textbf{Automated Evaluation}
We compute the perplexity of all models on the test set.
To evaluate model generations, we generate 200 responses from each model for 40 sampled contexts, and compare all model responses to the 10 gold generations for each context using our proposed matching algorithm (\S \ref{sec:matching}), using BLEU-4 and ROUGE-L F1 \citep{lin2004rouge} as scoring functions.


\vspace*{2mm}
\noindent
\textbf{Human Evaluation}
We have annotators evaluate the models' generations.
We give participants the prompt, conversation history, and a set of 10 model-generated responses, and ask them asked to rate three aspects of the responses (Diversity, Relevance, and Fluency) on a 5-point Likert scale.
We collect 1000 ratings (of 10 generations each) per model.


\begin{table}[tp]
\centering
\resizebox{\columnwidth}{!}
{\begin{tabular}{@{}lccc@{}}
\toprule
\textbf{Model} & \textbf{PPL} & \multicolumn{1}{p{3cm}}{\centering \textbf{Matching Metric (BLEU-4)}}  & \multicolumn{1}{p{3cm}}{\centering \textbf{Matching Metric (ROUGE-L f1)} } \\ \midrule
GPT-2 S & 51.63 &  0.06   &  3.62 \\
 GPT-2 M
  & 41.95  &  0.14  &    3.87  \\
  GPT-2 L &  35.67   &  0.16   &    4.03 \\
GPT-2 XL   &  33.91     &  0.14   &    4.04 \\ 
Dialog Tree Model                                                                  &    \textbf{13.21}    &  \textbf{0.28} &   \textbf{4.32} \\ \bottomrule
\end{tabular}}
\caption{
Test performance for Task 1 (Diverse Generation).
}
\label{tab:task0}
\end{table}

\begin{table}[tp]
\centering
\resizebox{\columnwidth}{!}
{\begin{tabular}{@{}lccc@{}}
\toprule
\textbf{Model} & \textbf{Diversity} & \multicolumn{1}{p{3cm}}{\centering \textbf{Relevance}}  & \multicolumn{1}{p{3cm}}{\centering \textbf{Fluency} } \\ \midrule
GPT-2 S & 3.91 &  2.05   &  2.81 \\
 GPT-2 M & 3.78  &  2.67  &    3.42  \\
  GPT-2 L &  3.68   &  2.86   &    3.58 \\
GPT-2 XL   &    3.92   &  3.02   &    3.81 \\ 
Dialog Tree Model                                                               &    3.93    &  \textbf{3.28} &   \textbf{4.17} \\
\midrule
Gold Responses  &    4.06   &  4.49 & 4.61 \\
\bottomrule
\end{tabular}}
\caption{
Human evaluation results (5-point Likert scale) for the Diverse Generation Task (Task 1).
Bold numbers are a statistically significant improvement over the next highest result ($t$-test, $p<0.001$).
\textit{Takeaway:} A fine-tuned model is able to improve upon Relevance and Fluency, while maintaining Diversity; still, there is significant room for improvement to human level, especially in Relevance.
}
\label{tab:task0_human_eval_scoring}
\end{table}

\begin{table*}[t]
\centering
\resizebox{2\columnwidth}{!}
{\begin{tabular}{@{}llrccccccccc@{}}
\toprule
 & &\textbf{Eval} & \multicolumn{8}{c}{\textbf{\textit{Emotion classifier accuracy}}} & \textbf{Emotion Acc.}\\ \cmidrule(r){4-11}
\textbf{Model} & \textbf{Setup}& \textbf{PPL} & \textbf{Average}  & \textbf{Neutral} & \textbf{Joy} & \textbf{Sadness} & \textbf{Fear} & \textbf{Anger} & \textbf{Surprise} & \textbf{Disgust} & \textbf{Human Eval.} \\ \midrule
\textbf{GPT-2 M} &&63.41& 0.14& 0.66&0.09 &0.07 & 0.03 &0.04 & 0.08 & 0.05 & 1.59 \\ 
\textbf{Dialog Tree Model}& &\textbf{7.03} &0.14 & 0.65& 0.08&0.05 & 0.04 & 0.05&0.08&0.04 & 1.64 \\ 
\textbf{Retrieval} & most likely &16.88 &0.14 & 0.67&0.11 &0.06 & 0.03 & 0.06&0.06&0.02 & 1.63 \\ 
& w/ emotion $e$&17.64 & \textit{1.00}&\textit{1.00} & \textit{1.00}&\textit{1.00} & \textit{1.00} & \textit{1.00}& \textit{1.00} &\textit{1.00} & \textbf{2.13} \\
\textbf{Emotion-Aware Model} & normal &7.53 &0.51 & \textbf{0.84}&0.60 & 0.38& 0.40 & 0.39& 0.56 &\textbf{0.42} & \textbf{2.18} \\
&oversampled&8.33 &\textbf{0.54} &0.76 &\textbf{0.64} & \textbf{0.42}& \textbf{0.48} &\textbf{0.48}& \textbf{0.60} & 0.40 & \textbf{2.11} \\ \bottomrule
\end{tabular}}
\caption{
Test set performance for Emotion-Conditioned Generation task. For human evaluation (3-point Likert scale, rightmost column), bolded numbers are a statistically significant improvement from others ($t$-test, $p<0.001$).
\textit{Takeaway:} The \emoModelName~successfully balances generation coherency (as measured by \textit{Eval PPL}) with adhering to the specified emotion. 
}
\label{tab:task1}
\end{table*}

\vspace*{2mm}
\noindent
\textbf{Results}
Tables \ref{tab:task0} and \ref{tab:task0_human_eval_scoring} show the results for the Diverse Generation Task, comparing the Dialog Tree Model (a finetuned GPT-2 M) against all out-of-the-box pretrained GPT-2 sizes.
We observe an expected improvement on all metrics by carefully finetuning the model on the tree-structured corpus.
In human evaluation (Table \ref{tab:task0_human_eval_scoring}), the Dialog Tree Model significantly improves Relevance and Fluency, but ties at response Diversity.
While it is initially surprising that a tree-trained model is not more diverse, the results are intuitive: the \textit{standalone diversity} of generations is strongly influenced by the sampling procedure \cite{holtzman2019curious}; and maximizing diversity, \textit{as well as} measures of quality, indicate good generations \cite{Caccia2020Language}.
Promisingly, the matching metric (Table \ref{tab:task0}) captures the strongest (fluent and relevant) model, even when it exhibits the same diversity as others.

We present these results as baseline numbers for the diverse generation task, and leave efforts to explicitly target even greater generation diversity to future.

\subsection{Task 2: Emotion-Conditioned Generation}

In this task, a model is given the complete context and desired emotion ($u_0, ..., u_{i-1}, e$), and should generate a response $u_i$ with emotion $e$.
We introduce the Emotion-Aware Model for this task (\S\ref{sec:emotion-aware-model}).

\vspace*{2mm}
\noindent
\textbf{Baselines}
In addition to GPT-2 Medium (both out-of-the-box and fine-tuned), we also use retrieval baselines. 
The first retrieval baseline finds the training set context with the highest similarity.
We average GloVe embeddings \cite{pennington2014glove}\footnote{Specifically: 40B, 300d GloVe vectors.} of all words in the contexts and score with cosine similarity.
The second retrieval baseline performs the same similarity calculation, but is restricted to choosing from utterances with the desired emotion $e$.

\vspace*{2mm}
\noindent
\textbf{Automated Evaluation}
We sample 20 contexts from each prompt in the test set, and generate 7 responses per prompt, one for each emotion $e \in E$.
Model generations are evaluated both on whether they are classified as emotion $e$ (emotion acc.), and with a PPL score given to them by an emotion-agnostic LM fine-tuned on the dataset (coherence).

\vspace*{2mm}
\noindent
\textbf{Human Evaluation}
We also measure human ratings of emotion accuracy on a 3-point Likert scale for 1000 generations per model.
Whereas in Task 1, workers saw a set of 10 generations (to evaluate diversity), here workers rate each generation individually.
\vspace*{2mm}
\noindent
\textbf{Results}
Table \ref{tab:task1} shows the experimental results for the Emotion-Conditioned Generation task.
While the retrieval baseline is trivially able to retrieve a response with the correct emotion, it suffers by having significantly worse perplexity as judged by the independent fine-tuned LM, and lower relevance and fluency (by humans).
The Emotion-Aware model is able to balance coherent replies (low Eval. PPL) while accurately generating responses with the conditioned emotion (both automatic and human-evaluated).
Training the Emotion-Aware Model on an oversampled subset leads to a slight improvement in emotion adherence  at the cost of higher perplexity. 

\subsection{Task 3: Theory of Mind Generation}

In this task, models generate a response $u_i$ that will lead to a conversation partner to a response $u_{i+1}$ with emotion $e$.
We propose the Lookahead Model for this task (\S\ref{sec:lookahead-model}).


\vspace*{2mm}
\noindent
\textbf{Baselines}
We run all of the baselines and models from Task 2 as baselines for this task.
Though the goal is different, it is intuitively plausible that simply generating a response $u_i$ with emotion $e$ may cause a conversation partner to \textit{mirror} that same emotion $e$ in their reply $u_{i+1}$.

In addition, we create an emotion transition matrix $T$ from the training data that contains which emotions are likely to lead to each other.
We denote $T[\rightarrow e]$ the emotion of utterance $u_i$ most likely to \textit{lead} to emotion $e$ for the reply $u_{i+1}$.
We use $T[\rightarrow e]$ as additional variants both for the retrieval baseline, and for conditioning the Emotion-Aware Model.


\vspace*{2mm}
\noindent
\textbf{Evaluation}
As with Task 2, we sample 20 contexts from each prompt in the test set, and generate 7 responses per prompt, one corresponding to each emotion $e \in E$.
Then, for each response, we have a fine-tuned LM as a conversation partner continue the conversation by generating a $u_{i+1}$, which the emotion classifier then assesses for a match to $e$.
The fine-tuned LM also scores the perplexity of the replies $u_i$.
We report the average of 5 runs for all numbers to mitigate generation randomness.

\begin{table*}[tph]
\centering
\resizebox{2\columnwidth}{!}
{\begin{tabular}{@{}llrccccccccc@{}}
\toprule
                                                                                         &     &  \textbf{Eval}   & \multicolumn{9}{c}{\textbf{\textit{Emotion classifier accuracy}}}                                             \\ \cmidrule(r){4-12}
\textbf{Model} & \textbf{Setup}                                                                                    & \textbf{PPL} & \textbf{Average} & \textbf{No-neutral avg.} & \textbf{Neutral} & \textbf{Joy} & \textbf{Sadness} & \textbf{Fear} & \textbf{Anger} & \textbf{Surprise} & \textbf{Disgust} \\ \midrule
GPT-2 M                                                                 &       &  67.79   &   0.14      &         0.04            &       0.74  &  0.07   &    0.05     &   0.02   &    0.03   &     0.06     &     0.04    \\ 
Dialog Tree Model                                                                  &       &  7.11   &    0.15     &      0.05               &   0.77      &   0.09  &    0.05     &   0.02   &   0.03    &    0.05      &    0.03     \\ 
Retrieval   &  most likely   &  16.72   &    0.14     &     0.04                &    0.74     &  0.06   &    0.05     &   0.02   &   0.04    &       0.06  &    0.03     \\ 
  & w/ $e$  &  18.33   &    \textbf{0.16}     &         \textbf{0.06}            &      0.73   &   \textbf{0.13}  &    0.06     &   \textbf{0.04}   &   0.05    &     0.04     &    \textbf{0.07}     \\  & w/ T{[}$\rightarrow e${]}&  17.90   &  0.15       &      0.05               &   0.77      &  0.08   &     0.04    &   0.02   &   0.04    &     \textbf{0.07}     &    0.04     \\
   Emotion-Aware Model         &   w/ $e$   &  7.68   &    \textbf{0.16}    &   \textbf{0.06}                  &    0.74     &  0.11   &   0.05      &   0.02   &      \textbf{0.06}  &     \textbf{0.07}     &     0.03    \\ 
   &   w/ T{[}$\rightarrow e${]}   &  7.21   &    0.14     &            0.04         &   0.73      &  0.09   &    0.04     &  0.03   &   0.03    &    0.05      &    0.03     \\
        &   w/ $e$ + oversampled   &  8.47   &    0.15     &   \textbf{0.06}                  &    0.75     &  0.10   &   \textbf{0.07}      &   0.03   &      0.04  &     0.04     &     0.04    \\ 
   &   w/ T{[}$\rightarrow e${]} + oversampled  &  8.50   &    0.15     &            0.04         &   \textbf{0.78}      &  0.07   &    0.05     &  0.03   &   0.03    &    0.05      &    0.03     \\
\modelName~                                                                         &   normal    &  \textbf{6.62}   &   0.15      &     0.05                &       0.76  &  0.08   &   0.05      &  0.03    &   0.03    &     0.05     &    0.03     \\
&  oversampled    &  7.30   &    0.14     &      0.04               &   0.75      &  0.06   &   0.05      &   0.03   &    0.03   &    0.06      &  0.02       \\ \midrule
Oracle &    &  13.90 &  0.17   &    0.07     &      0.78               &   0.13      &  0.09   &   0.04      &   0.04   &    0.09   &    0.03           \\ \bottomrule
\end{tabular}}
\caption{
Test set performance for Theory of Mind Generation task.
All numbers are averaged over 5 runs.
\textit{Takeaway:} While the \modelName~achieves the lowest perplexity, all models struggle to reliably influence the emotion of the next prediction. Approaches that mirror the desired emotion $e$ reliably perform best, indicating this may be a hard-to-beat strategy.
}
\label{tab:task2}
\end{table*}
\begin{figure*}[ht]

\begin{center}
\includegraphics[width=0.99\linewidth]{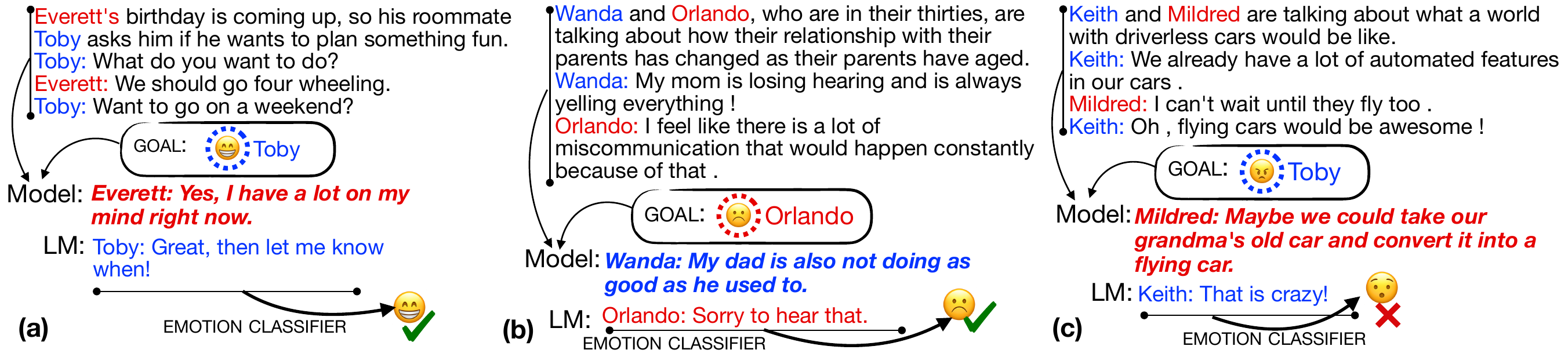}
\end{center}
\caption{
Three generation examples from the Lookahead Model for the Theory of Mind Task. 
}
\label{fig:lookahead_examples}
\end{figure*}

\vspace*{2mm}
\noindent
\textbf{Results}
Table~\ref{tab:task2} presents the experimental results for the Theory of Mind task.
Unlike the previous two tasks, this task does not present a clear winner.
The Lookahead Model, whose training scheme most closely mimics the task, achieves the lowest evaluated perplexity according to the external model (6.62).
However, it struggles as much as emotion-agnostic baselines to influence emotion states of its conversation partner (0.14).
The strongest models at this task (0.16) are \textit{mirroring} approaches: the retrieval baseline and Emotion-Aware Model, both using the conversation partner's target emotion $e$ as their parameter.

Even the oracle---which uses responses from the validation set known to lead to the desired emotion---is not far ahead (0.17).
This results hints at the significant challenge of this task, which we discuss in the next section.

\subsection{Qualitative Analysis}

Figure \ref{fig:lookahead_examples} gives three examples from the Lookahead Model's output on the dev set, which illustrate some of the main response categories.
In (a), the model performs what appears to be relatively complex maneuver: by outputting a relatively melancholy response (\textit{Yes, I have a lot on my mind right now}), the conversation partner appears to ``cheer them up'' by being excited about their plans.
Example (b) shows \textit{mirroring}: the model says something sad, and the LM replies with something that is also sad.
The approach served other models well at directing conversations.
Finally, example (c) illustrates a common problem: it can be difficult to consistently direct an LM conversation partner.
While a language model fine-tuned on our dataset is exposed to a variety of responses for many contexts, this same diversity can lead to challenges reliably influencing the behavior of the model.
Investigating the role of cooperative agents \cite{lewis2017deal} in a diverse dialog setting may prove an interesting avenue of future work.
\section{Related Work}

\noindent
\textbf{Dialog Datasets and Personas}
Previous dialog datasets have been created from scraped conversations of social media, such as Twitter \cite{ritter2010unsupervised} and Weibo \cite{shang2015neural}.
Other recent work has proposed that Chit-chat models become more engaging when endowed with a consistent ``persona'' \citep{li2016persona, zhang2018personalizing, shuster2019engaging}.
Several datasets have taken advantage of this idea and included personas for the participants.
\textsc{Persona-Chat} \citep{zhang2018personalizing} dialogues revolve around fictional speakers' personas.
%
%
Our dataset contains implicit personas from the conversation prompts, rather than explicit attributes assigned to each character.

\vspace*{2mm}
\noindent
\textbf{Emotion Recognition in Conversations}
The appearance of more large-scale conversational datasets in recent years has coincided with research focused on emotion recognition
\citep{abdul2017emonet, chen2018emotionlines, zhou2017emotional, poria2018meld}.
RNN-based networks enjoyed widespread use as the foundation for such models
%
%
\citep{abdul2017emonet, majumder2019dialoguernn}.
\citet{hazarika2018conversational} introduced a conversational memory network (CMN) for multimodal emotion recognition.
%
%
\citet{hazarika2019emotion} proposed \textit{TL-ERC}: transfer learning from a neural dialogue generation model for emotion recognition.
\citet{li2020multi} proposed a multi-task learning network with the assistance of speaker identification for emotion recognition.

For dialogue generation, \citet{zhou2017mojitalk, zhou2017emotional, wang2018sentigan, li2019reinforcement} generated responses with corresponding emotions using Conditional Variational Autoencoders (CVAEs) \citep{sohn2015learning}, GANs \citep{goodfellow2014generative}, and Deep RL.
The core of our approach, like other recent work, is strategically finetuning large pretraied LMs \citep{radford2019language}.




\section{Conclusion}

We present the large-scale, highly branching conversational \datasetName.
In studying this dataset, we propose a multi-reference scoring algorithm, as well as models that look ahead to optimize for future conversation states.
We hope the introduction of our data and methods will spur more exciting work in multi-path conversational dialog.

\bibliography{aaai21.bib}
\end{document}